\documentclass[
]{ceurart}

\usepackage{listings}
\lstdefinelanguage{SPARQL}{
  basicstyle=\small\ttfamily,
  backgroundcolor=\color{LightGray},
  columns=fullflexible,
  breaklines=false,
  sensitive=true,
  frame=bt,
  aboveskip=1em,
  belowskip=1em,
  xleftmargin=.5em,
  xrightmargin=.5em,
  framexleftmargin=.5em,
  framextopmargin=.5em,
  framexbottommargin=.5em,
  framexrightmargin=.5em,
  tabsize = 2,
  showstringspaces=false,
  morecomment=[l][\color{gray}]{\#},       
  morecomment=[n][\color{blue}]{<http}{>}, 
  morestring=[b][\color{OliveGreen}]{\"},  
  keywordsprefix=?,
  classoffset=0,
  keywordstyle=\color{Sepia},
  morekeywords={},
  classoffset=1,
  keywordstyle=\color{Purple},
  morekeywords={rdf,rdfs,owl,xsd,purl},
  classoffset=2,
  keywordstyle=\color{MidnightBlue},
  morekeywords={
    SELECT,CONSTRUCT,DESCRIBE,ASK,WHERE,FROM,NAMED,PREFIX,BASE,OPTIONAL,
    FILTER,GRAPH,LIMIT,OFFSET,SERVICE,UNION,EXISTS,NOT,BINDINGS,MINUS,a
  }
}

\begin{document}

\copyrightyear{2024}
\copyrightclause{Copyright for this paper by its authors.
  Use permitted under Creative Commons License Attribution 4.0
  International (CC BY 4.0).}

\conference{SEMANTiCS 2024: 20th International Conference on Semantic Systems, September 17--19, 2024, Amsterdam, The Netherlands}
\title{NFDI4DSO: Towards a BFO Compliant Ontology for Data Science}


\author[1,2]{Genet Asefa Gesese}[%
orcid=0000-0003-3807-7145,
email=genet-asefa.gesese@fiz-kalrsruhe.de,
url=https://tinyurl.com/3cx37b9x,
]
\cormark[1]
\address[1]{FIZ Karlsruhe, Leibniz Institute for Information Infrastructure, Germany }
\address[2]{Karlsruhe Institute of Technology, KIT, Germany}

\author[1,2]{Jörg Waitelonis}[%
orcid=0000-0001-7192-7143,
email=Joerg.Waitelonis@fiz-Karlsruhe.de,
url=https://shorturl.at/UwDND,
]
\author[3]{Zongxiong Chen}[%
orcid=0000-0003-2452-0572,
email=zongxiong.chen@fokus.fraunhofer.de,
]
\address[3]{Fraunhofer FOKUS, Berlin, Germany}

\author[3]{Sonja Schimmler}[%
orcid=0000-0002-8786-7250,
email=sonja.schimmler@fokus.fraunhofer.de,
url=https://www.fokus.fraunhofer.de/009785fd54551039,
]

\author[1,2]{Harald Sack}[%
orcid=0000-0001-7069-9804,
email=harald.sack@fiz-kalrsruhe.de,
url=https://www.aifb.kit.edu/web/Harald\_Sack,
]

\cortext[1]{Corresponding author.}

\begin{abstract}
The NFDI4DataScience (NFDI4DS) project aims to enhance the accessibility and interoperability of research data within Data Science (DS) and Artificial Intelligence (AI) by connecting digital artifacts and ensuring they adhere to FAIR (Findable, Accessible, Interoperable, and Reusable) principles. To this end, this poster introduces the NFDI4DS Ontology, which describes resources in DS and AI and models the structure of the NFDI4DS consortium.   Built upon the NFDICore ontology and mapped to the Basic Formal Ontology (BFO), this ontology serves as the foundation for the NFDI4DS knowledge graph currently under development.

\end{abstract}


\maketitle

\section{Introduction}
The German National Research Data Infrastructure (NFDI)\footnote{\url{https://www.nfdi.de/}} is a non-profit association founded to coordinate the activities for establishing a national research data infrastructure. It comprises 26~consortia spanning a wide range of scientific disciplines, from cultural sciences, social sciences, humanities and engineering to life sciences and natural sciences.  The NFDI consortia share common goals and concepts, such as their members, structure, data repositories, and services~\cite{sack2023knowledge}. To enhance interoperability across these consortia, the NFDICore ontology\footnote{\url{https://ise-fizkarlsruhe.github.io/nfdicore/2.0.0/}} has been developed. It acts as a mid-level ontology for representing metadata related to NFDI resources, including individuals, organizations, projects, data portals, and more.  NFDICore provides mappings to a broad range of standards across different domains, such as the Basic Formal Ontology (BFO)~\cite{otte2022bfo} and Schema.org~\cite{guha2016schema} to advance knowledge representation, data exchange, and collaboration across diverse domains. To address domain-specific research questions for each consortium, NFDICore follows a modular architecture. Examples for modular extensions include the NFDI4Culture ontology module CTO\footnote{\url{https://gitlab.rlp.net/adwmainz/nfdi4culture/knowledge-graph/culture-ontology}}\cite{tietz2023damalos} and the NFDI-MatWerk ontology module MWO\footnote{\url{https://git.rwth-aachen.de/nfdi-matwerk/ta-oms/mwo}}, which are specifically designed for the cultural heritage and materials science domains, respectively. In this paper, we present an ontology named NFDI4DSO for the data science domain as a domain-specific modular extension of NFDICore.

NFDI4DataScience (NFDI4DS)\footnote{\url{https://www.nfdi4datascience.de/}} is one of the NFDI consortia and its project aims to enhance the accessibility and interoperability of research data in the domain of Data Science (DS) and Artificial Intelligence (AI). Data Science (DS) is a multidisciplinary field combining different aspects of mathematics, statistics, computer science, and domain-specific knowledge to extract meaningful insights from diverse data sources. DS and Artificial Intelligence (AI) involve various artifacts, e.g., datasets, models, ontologies, code repositories, execution platforms, repositories, etc. The project achieves this by linking digital artifacts and ensuring their FAIR (Findable, Accessible, Interoperable, and Reusable) accessibility, thereby fostering collaboration across various DS and AI platforms. To this end, the NFDI4DS Ontology (NFDI4DSO) is built. 

\section{The NFDI4DataScience Ontology (NFDI4DSO)}
As mentioned earlier, NFDI4DSO is created in a modular fashion, building upon NFDICore. Similar to  NFDICore, the NFDI4DSO ontology is developed using a bottom-up, iterative, user-centered approach. NFDICore comprises 51~classes, 55~object properties, 8~data properties, 18~annotation properties, and 5~SWRL rules~\cite{horrocks2004swrl} (for details refer to NFDICore documentation\footnote{\url{https://ise-fizkarlsruhe.github.io/nfdicore/}}). In NFDI4DSO, in addition to what is provided in NFDICore,  42~classes,  38~object properties, 9~data properties, 
and  8~SWRL rules are added.  The NFDI4DSO ontology not only describes various data science artifacts but also provides information about the resources of the NFDI4DS Consortium, such as personas, consortium members, spokespersons, and task area leads. AS in NFDICore, the classes introduced in NFDI4DSO are also mapped to the top-level ontology BFO and also other ontologies such as schema.org, the FaBiO ontology~\cite{peroni2012fabio}, and the Conference Ontology\footnote{\url{http://www.scholarlydata.org/ontology/doc/\#toc}}.

\begin{figure}
    \centering
    \includegraphics[scale=0.4]{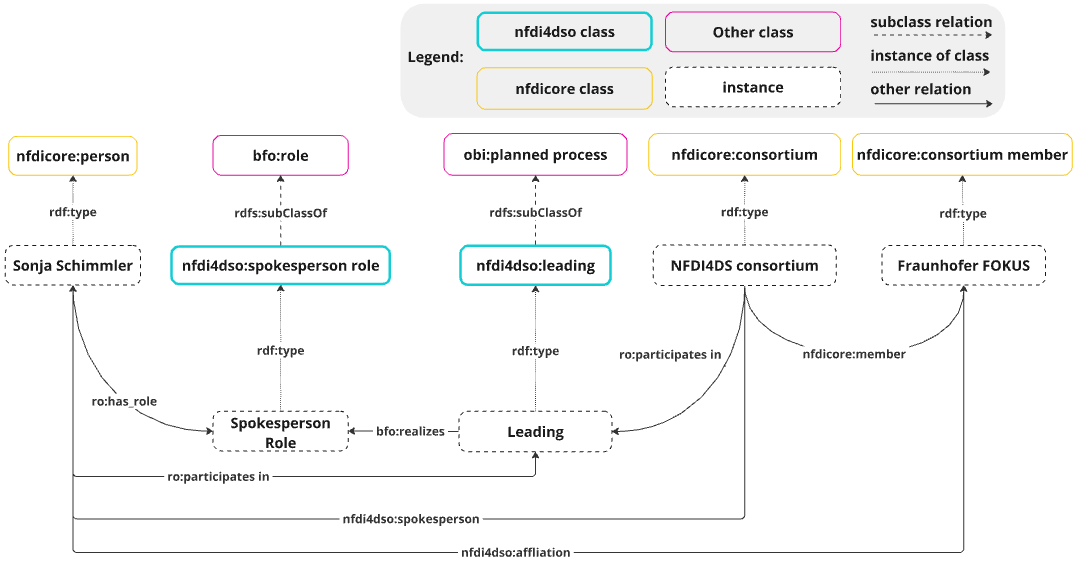}
    \caption{Example of representing roles where the prefixes ro and obi represent http://purl.obolibrary.org/obo/ro.owl and http://purl.obolibrary.org/obo/obi.owl ontologies, respectively.}
    \label{fig:role}
\end{figure}

NFDI4DSO contains various kinds of classes such as processes, roles, and independent continuants.  For instance, Figure~\ref{fig:role} depicts how NFDI4DSO represents the relationship between the independent continuant \textit{nfdi4dso:SonjaSchimmler} and her spokesperson role \textit{nfdi4dso:SpokespersonRole} by mapping it to BFO. By using roles and processes, NFDI4DSO enables a detailed representation of the relationship between different entities enhancing the ontology’s level of expressivity. On the other hand, to support easier integration and use of less complex relations, shortcuts are also introduced to simplify the ontology by implementing easy-to-use direct shortcut properties, which can be expanded to fully-fledged BFO-compliant complex path expressions. For instance, in Figure~\ref{fig:role}, the shortcut relation \textit{nfdi4dso:spokesperson} is provided and its corresponding  SWRL\footnote{\url{https://ise-fizkarlsruhe.github.io/NFDI4DS-Ontology/\#d4e7620}} rule is given below.

\textit{Person(?p)\ $\wedge$ \ Consortium(?c)\ $\wedge$ \  SpokespersonRole(?sr)\ $\wedge$ \  Leading(?l)\ $\wedge$ \  participates in(?p, ?l)\ $\wedge$ \  participates in(?c, ?l)\ $\wedge$ \  has role(?p, ?sr)\ $\wedge$ \  realised in(?sr, ?l)\ $\rightarrow$ spokesperson(?c, ?p)}

\paragraph{Ontology Implementation}
The Protégé ontology editor \footnote{https://protege.stanford.edu/} for the OWL-based formalization of terminological knowledge has been used to develop and implement NFDI4DSO. Widoco\footnote{https://github.com/dgarijo/Widoco} has been used to create an enriched and customized documentation of the ontology automatically.  The stable version of the ontology NFDI4DSO v1.0.0 is available at \url{https://github.com/ISE-FIZKarlsruhe/NFDI4DS-Ontology/tree/main} and the latest development version is at \url{https://github.com/ISE-FIZKarlsruhe/NFDI4DS-Ontology/tree/develop-1.0.1}.

\section{NFDI4DSO in Use}

\begin{figure}
    \centering
    \includegraphics[scale=0.35]{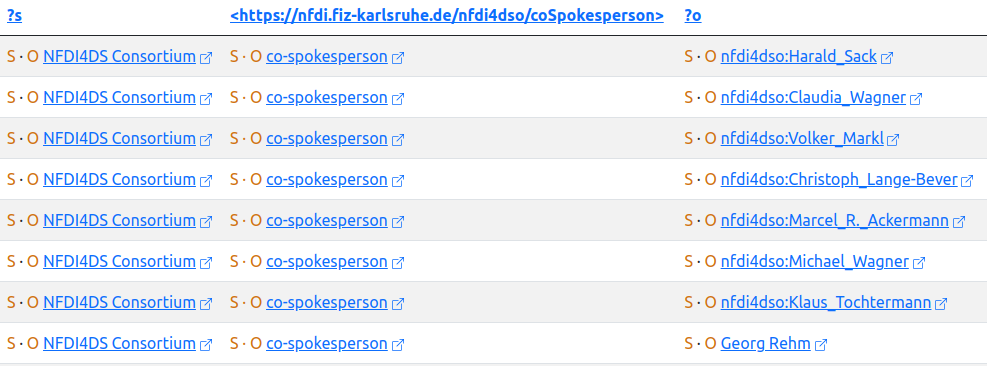}
    \caption{A screenshot of part of the SHMARQL interface with the list of NFDI4DS co-spokespersons (refer to \url{https://shorturl.at/eNb5e} to navigate it fully.)}
    \label{fig:shmarql}
\end{figure}

\begin{figure}
    \centering
    \includegraphics[scale=0.4]{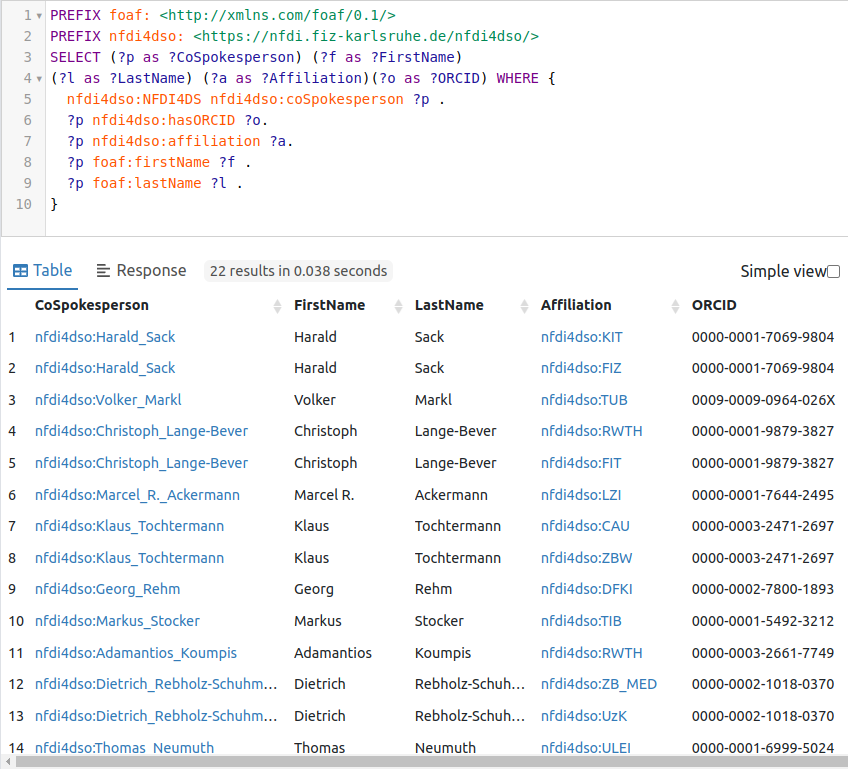}
    \caption{An example SPARQL query to provide a list of the co-spokespersons of the NFDI4DS Consortium. (It possible to query it live at: \url{https://nfdi.fiz-karlsruhe.de/4ds/sparql})}
    \label{fig:sparql}
\end{figure}

The NFDI4DSO is designed to form the foundation of the NFDI4DS Knowledge Graph (NFDI4DS-KG), which is currently under development. The NFDI4DS-KG consists of two main components: the Research Information Graph (RIG) and the Research Data Graph (RDG). RIG includes metadata about the NFDI4DS consortium’s resources, persons, and organizations, while the RDG encompasses content-related index data from the consortium’s heterogeneous data sources. RIG serves as the backend for the NFDI4DS web portal, facilitating interactive access and management of this data. Both RIG and RDG will be accessible and searchable via the NFDI4DS Registry platform. Additionally, the NFDI4DS consortium plans to collaborate with other NFDI consortia to further integrate domain-specific knowledge into the RDG seamlessly. Currently, the first version of the NFDI4DS-KG\footnote{\url{https://nfdi.fiz-karlsruhe.de/4ds/sparql}, \url{https://nfdi.fiz-karlsruhe.de/4ds/shmarql}} with RIG is publicly available.  For example, to view the list of co-spokespersons of the NFDI4DS Consortium, you can either navigate through the data using SHMARQL\footnote{\url{https://shorturl.at/eNb5e}}, as depicted in Figure~\ref{fig:shmarql} or query it using SPARQL, as shown in Figure~\ref{fig:sparql}.

\section{Conclusion and Future Work}
This paper presents the NFDI4DS Ontology and its use for the  NFDI4DS-KG that is currently under-development. The ontology facilitates the representation and interoperability of data science artifacts within and outside of NFDI4DS.  NFDI4DSO is built on top of the NFDICore ontology and mapped to BFO and other ontologies. In the future, there is a plan to perform extensive ontology evaluation using competency questions based on the persona definitions from the NFDI4DS consortium.

\begin{acknowledgments}
   This publication was written by the NFDI consortium NFDI4DataScience in the context of the work of the association German National Research Data Infrastructure (NFDI) e.V.. NFDI is financed by the Federal Republic of Germany and the 16 federal states and funded by the Federal Ministry of Education and Research (BMBF) – funding code M532701 / the Deutsche Forschungsgemeinschaft (DFG, German Research Foundation) - project number NFDI4DataScience (460234259).
\end{acknowledgments}

\bibliography{biblio}

\appendix

\end{document}